\documentclass[a4paper]{easychair}
\usepackage[latin1]{inputenc}
\usepackage[english]{babel}
\usepackage{epic}
\usepackage{epstopdf}
\usepackage{graphicx}
\usepackage{url}
\usepackage{mathptmx}
\usepackage{indentfirst}
\usepackage{amssymb}
\usepackage{amsmath}
\usepackage{textcomp}
\input{qobitree}

\newcommand\Land{\&^\Pi}

\newcommand\type[1]{^{#1}}
\newcommand\et\land
\newcommand\fl{\rightarrow}

\newcommand\ttt{\mathbf{t}}
\newcommand\eee{\mathbf{e}}

\usepackage{gb4e} 
\begin{document}

\title{Advances in the Logical Representation\\ of Lexical Semantics
}


\titlerunning{Advances in the Logical Representation of Lexical Semantics}

\author{Bruno Mery\inst{1} \and Christian Retoré\inst{2}}

\institute{Université de Bordeaux, France\\
		\email{bruno.mery@me.com}
	\and
		IRIT-C.N.R.S., Université de Bordeaux, France\\
		\email{christian.retore@labri.fr}
}

\authorrunning{Mery and Retoré}

\maketitle

\begin{abstract}
The integration of lexical semantics and pragmatics in the analysis of the meaning of natural language has prompted changes to the global framework derived from Montague. In those works, the original lexicon, in which words were assigned an atomic type of a single-sorted logic, has been replaced by a set of many-facetted lexical items that can compose their meaning with salient contextual properties using a rich typing system as a guide.\par
Having related our proposal for such an expanded framework $\Lambda TY_n$, we present some recent advances in the logical formalisms associated, including constraints on lexical transformations and polymorphic quantifiers, and ongoing discussions and research on the granularity of the type system and the limits of transitivity.\par
\end{abstract}

\section{Introduction}
Our general purpose is to update the framework colloquially known as ``Montague Grammar'', (based on principles described in \cite{montague-proper}) used in formal linguistics to provide an analysis of the syntax, semantics and interpretation of the human language, with data pertaining to lexical semantics according to \cite{gen-lexicon}.\par
Our main principles are the following:
\begin{itemize}
\item The semantics should retain its compositional aspects.
\item The lexical semantics should allow to distinguish between acceptable and unacceptable predications.
\item The mechanism should allow some degree of flexibility for different contexts.
\item The framework should be easy to define, describe and implement.
\end{itemize}

The simplest example of situation considered acceptable by standard Montagovian analysis is selection restriction that prescribes\footnote{We do not say that this sentence should never receive a semantical analysis. There are some contexts that can provide a suitable parse, and even in the absence of those, something may be derived (if only that there must exist some unprovided context that makes sense of the utterance). See Section \ref{infelicity}, p.~\pageref{infelicity} for more details.}, e.g.:

\begin{exe}
\ex * The chair barked.
\end{exe}

Restriction of selection, polysemy and more elaborate phenomena including the creative use of words and co-predicative statements have prompted the inception of the theory known as the Generative Lexicon, detailed in \cite{gen-lexicon}. In \cite{mery-jolli} and other related works, we have elaborated a Montague-compatible framework corresponding to that theory and the above principles. It has proved useful in the study of several phenomena, including deverbal nouns (\cite{real-deverbal}), the virtual traveler (\cite{moot-traveler}), etc. It is also currently partially implemented in a version compatible with $\lambda$-DRT.\par
Our goal here is to examine some of the mechanisms introduced by our framework, such as constraints on the availability of meanings and the modelling of articles with quantifiers, and to discuss several points pertaining to the type system and its granularity.

\subsection{Montague Grammar}
\label{montague}
Richard Montague, in \cite{montague-proper} and other related works, expressed the idea that any human language could be considered as a formal language, and receive a logical analysis that could be interpreted in terms of truth. He had a (rather Chomskyian) universal aim, resulting in the analysis process that would come to be called Montague Grammar, and was largely adopted after his death.\par
An abridged version of this process is the following:
\begin{enumerate}
\item The \emph{syntax} of the sentence is analysed using an appropriate formalism, such as categorial grammars. This provides a binary syntactic tree that indicates, minimally, at each node which subtree (the predicate) applies to which other (the argument).
\item A \emph{lexicon} provides a semantic term that corresponds to every word, for example as a simply-typed $\lambda$-term. The sentence is analysed as a logic formula by associating to each leaf of the syntactic tree (individual words) their corresponding term, and by using the tree to apply each argument to its predicate.
\item The formula can then be computed by $\beta$-reduction and interpreted, for example, as a truth value.
\end{enumerate}
Two common uses of the Montagovian analysis are:
\begin{itemize}
\item Using a lexicon with two types, $\eee$ for all entities and $\ttt$ for truth values; formulae are of type $\ttt$ and receive a boolean interpretation.
\item Using the above lexicon with the additional type $\mathbf{s}$, corresponding to indices for possible worlds; the interpretation is the set of indices satisfying the formula, in Kripke semantics, and can be combined with modal logic.
\end{itemize}

This is an example of a lexicon of the former kind.

\begin{figure}[!h]
\begin{center} 
\begin{tabular}{ll} \hline 
\textbf{word} &  \textbf{\itshape semantic type $u^*$}\\ 
& \textbf{\itshape  semantics~: $\lambda$-term of type $u^*$}\\ 
&  {\itshape  $x\type{v}$ the variable or constant $x$ 
is of type $v$}\\ \hline 
\textit{some} 
& $(e\fl t)\fl ((e\fl t) \fl t)$\\ 
& $\lambda P\type{e\fl t}\  \lambda Q\type{e\fl t}\  
(\exists\type{(e\fl t)\fl t}\  (\lambda x\type{e}  (\et\type{t\fl (t\fl t)} (P\ x) (Q\ x))))$ \\  \hline 
\textit{club}  & $e\fl t$\\ 
& $\lambda x\type{e} (\texttt{club}\type{e\fl t}\  x)$\\  \hline 
\textit{defeated} & $e\fl (e \fl t)$\\ 
& $\lambda y\type{e}\  \lambda x\type{e}\  ((\texttt{defeat}\type{e \fl (e \fl t)}\  x)  y)$ \\  \hline 
\textit{Leeds} &$e$ \\ &  Leeds 
\end{tabular}
\end{center} 
\caption{A simple semantic lexicon} 
\label{semanticlexicon}
\end{figure}

Considering the sentence
\begin{exe}
\ex Some club defeated Leeds
\end{exe}
Assuming the underlying syntactic structure is 
\begin{center}
(some\ (club)) (defeated\ Leeds) 
\end{center}
The analysis of the semantics is thus:
$$
\begin{array}{c} 
\Big(\big(\lambda P\type{e\fl t}\ \lambda Q\type{e\fl t}\  (\exists\type{(e\fl t)\fl t}\  (\lambda x\type{e}  (\et (P\ x) (Q\ x))))\big)
\big(\lambda x\type{e} (\texttt{club}\type{e\fl t}\  x)\big)\Big) \\ 
\Big(
\big(\lambda y\type{e}\  \lambda x\type{e}\  ((\texttt{defeated}\type{e\fl (e\fl t)}\  x)  y)\big)\ Leeds\type{e}\Big)\\ 
\multicolumn{1}{c}{\downarrow \beta}\\ 
\big(\lambda Q\type{e\fl t}\  (\exists\type{(e\fl t)\fl t}\  (\lambda x\type{e}  (\et\type{t\fl (t\fl t)}  
(\texttt{club}\type{e\fl t}\  x) (Q\ x))))\big)\\ 
\big(\lambda x\type{e} \ ((\texttt{defeated}\type{e\fl (e \fl t)}\  x)  Leeds\type{e})\big)\\ 
\multicolumn{1}{c}{\downarrow \beta}\\ 
\big(\exists\type{(e\fl t)\fl t}\  (\lambda x\type{e}  (\et (\texttt{club}\type{e\fl t}\  x) ((\texttt{defeated}\type{e\fl (e\fl t)}\  x)  Leeds\type{e})))\big)
\end{array}
$$
The end result is the logical form of the sentence, that corresponds to the following formula in predicate calculus: $\exists x:e\  (\texttt{club}(x)\ \et\ \texttt{defeated}(x,Leeds))$.\par
\vspace*{1em}
The analysis in Montague Grammar is thus comprised of \emph{two} levels of logical analysis:
\begin{itemize}
\item The logic of assembly (glue logic, compositional calculus). Here, simply-typed $\lambda$-calculus with base types $e$ and $t$. The terms provide a proof to the formula given by the types in intuitionist logic, à la Curry-Howard.
\item The logic of semantic representations. Here, higher-order predicate calculus. The resulting formula can be interpreted as a truth value.
\end{itemize}

Montague Grammar provides a semantic representation. It can be interpreted in terms of truth, possible worlds where the representation is true, etc. This provides an analysis of the possible realisations of a sentence. Our aim, conversely, is to refine the level of meaning assembly in order to take into account precisely \emph{what} is said, and \emph{how} this is said at the lexical level.

\subsection{Polysemy and the felicity of sentences}
Yeoshua Bar-Hillel famously used the following example in 1960 to illustrate the difficulties of Machine Translation:

\begin{exe}
\ex
	\begin{xlist}
	\ex \label{bip} The box was in the pen.
	\ex \label{pib} The pen was in the box.
	\end{xlist}
\end{exe}
In (\ref{bip}), \emph{pen} is a container; in (\ref{pib}), \emph{pen} is a simple tool. This should not come as a surprise to any linguist; the notion that words have sometimes radically different meaning that do not always get carried after translation is known as \emph{polysemy}. What it meant was that the use of an atomic lexicon for translation or other tasks of natural language processing was insufficient. Some additional mechanisms were needed that would allow to choose, in the view of the context, the correct meaning for every word; in his report, Bar-Hillel said that this required encyclopeadic knowledge, and that was \emph{surely utterly chimerical and hardly deserve[d] any further discussion}.\par
However, those examples are hardly the most complicated problem posed by polysemy. \emph{Pen} as a container and \emph{pen} as a tool are examples of \emph{contrastive ambiguity}, that is, of two words having completely different meanings but the same pronunciation and/or writing. There are various ways to deal with that ambiguity.\par
More problematic is the case of \emph{logical polysemy}, in which a word can express several meanings that are related logically, yet different. A \emph{bank} can be a geographical feature or a financial institution (this is contrastive ambiguity), but, in the latter case, it can variously refer to a building, a society or some unnamed person playing a specific role in that society.\par
Consider the following:
\begin{exe}
\ex \label{chair}* The chair barked.
\ex \label{bank} The bank killed my account.
\end{exe}
Not only do we want our framework to be able to tell that (\ref{chair}) is not acceptable in most contexts (because \emph{chair} is not a valid argument for the predicate \emph{to bark}) while (\ref{bank}) is (\emph{even though} \emph{bank} and \emph{account} are not associated with \emph{to kill} in a normal sense), we also want it to be able to tell us what process is at work in (\ref{bank}).
\section{Our Generative Framework}
\subsection{The Generative Lexicon}
In \cite{pust-genlex-article}, James Pustejovsky argued that logical polysemy is an essential feature of human languages, allowing a speaker to make the language evolve by the way of creative use of words. This means that the lexicon cannot be enumerative as additional word senses are added all the time; he proposed that the lexicon should contain enough information to generate for each word the correct meaning it would receive in context. As this represents a set of mechanisms for word sense generation rather that fixed meanings, he called it the Generative Lexicon.
As described in \cite{gen-lexicon}, the framework is Montagovian, with the following features:
\begin{description}
\item[Types in the terms] -- Every term is typed, as in Montague grammar, but using a rich typing system. The point here is that application is constrained, and some predicates may only apply to a specific kind of argument. Thus, the logic implicitly used by the generative lexicon is $TY_n$, the typed logic with $n$ sorts, which has $n+1$ types: $\ttt$ for truth values, and as many types as needed that replace the Montagovian $e$. Those types are not arbitrary, but constructed according a hierarchy of concepts under the form of an ontology.
\newpage
\item[Ontological hierarchy] -- The constraints in application are type-driven. Thus, types are an important mean to categorise terms in the systems. The organisation of types is hierarchical and ontological, as subtypes are understood to be compatible with constraints that would require their supertype. This inheritance hierarchy is based upon dual hyponymy-hyperonymy relations, with \emph{vegetable} being a subtype of \emph{food}, in itself a subtype of \emph{physical object}, and so on. The construct of such an ontology is a subject in itself for Pustejovsky. In \cite{pustejovsky-type}, he argues that another feature of the Generative Lexicon, Qualia, forms more robust guideline that existent ontological resources for lexical semantics (notably \emph{WordNet}). As such, the ontology proposed in GL remains an outline of a work in progress. 
\item[Qualia] -- The most distinguishing feature of the Generative Lexicon is the inclusion of the staple Aristotelician \emph{Qualia}. James Pustejovsky, citing \cite{moravcik-article}, applies this old philosophical concept to words and lexemes. And, while those concepts are insufficient to describe a physical phenomenon, they are indeed associated to \emph{the idea} that humans have about a phenomenon. They are ontological concepts, part of metaphysics, of our understanding of the world rather than of the world itself, and as such quite close to linguistic concepts.\par
Thus, in the Generative Lexicon, each lexical item comprises a Qualia structure, divided in four parts: the \emph{formal quale}, pertaining to intrinsic properties; the \emph{constitutive quake}, pertaining to components or materials; the \emph{agentive quale}, pertaining to the origin; and the \emph{telic quale}, pertaining to the purpose.\par
\end{description}
It would, when complete, be able to deal with phenomena such as:
\begin{description}
\item[Contrastive polysemy]: the ``accidental'' polysemy between, e.g., \emph{bar} as a place and as a physical object. The refined typing should be enough to distinguish between the two, as one would be typed $Location$ and the other $Physical$.
\item[Accommodation]: hyponymic relations, where a word is taken to be its ontological supertype, are included in the system. Thus, locutions such as \emph{my Honda} can be applied to predicates that require cars or physical objects.
\item[Qualia exploitation]: the four qualia are designed to be used in place of the word itself if required by context or sentence structure. Thus, a \emph{naïve article} should be parsed using the agentive quale because \emph{naïve} takes an agent as its argument. This particular case is a \emph{type coercion}: the predicates calls for a specific type which is incompatible with the one of the argument, but there is a transformation available to coerce the argument into the required type via qualia exploitation.
\item[Lexical transformation]: some other operations can be devised within the lexicon. \cite{copestake-lexical}, for instance, suggests \emph{grinding}, which can make available to the semantics the meaning of something that has undergone a destructive operation. \emph{Delicious chicken}, for example, indicates that the lexical item referred to as \emph{chicken} should not be considered as having an \emph{animal} type, but a \emph{food} type, as allowed by a grinding operation (specifically, cooking).
\item[Facettes]: some words can refer to different aspects, or \emph{facettes}, and have proved very difficult to model in the Generative Lexicon. They are the object of specific mechanisms. The canonical example is \emph{book}, as the word can refer to either a \emph{physical} object, made of leather, cardboard and paper, or its \emph{information} content that can be written or read.
\end{description}
\newpage
However, the Generative Lexicon was incomplete. The case study is linguistically impressive and well motivated, but the assembly language is not really described, and while every structure is detailed, a correct way to perform the desired derivation is simply assumed. There is no formal guide that can distinguish between the different cases, no algorithm that would choose the correct construction in an analysis. In fact, what is missing is an actual model for composition.\par
 As such, the Generative Lexicon was not directly suited for formal or computational implementation. It also famously lacked a comprehensive logical framework for complex types and co-predicative sentences. The author was aware of the latter fact and started working on a logically sound system in \cite{pustejovsky-metaphysics}, that would later be continued by Asher.\par
Several solutions have then been proposed, including ours (in \cite{mery-ndttg}, \cite{mery-montagl}, \cite{mery-jsm}, \cite{mery-jolli}), systems by Cooper (\cite{cooper-codygeqlic}), Asher (\cite{asher-fundamenta}, \cite{asher-webofwords}) and Luo (\cite{luo-dot}, \cite{luo-contextual}, \cite{luo-types}).
\subsection{The Logic and Calculus}
Our own proposal is based on the following elements:
\begin{itemize}
\item The description language (the logic used for final semantic representations) is the many-sorted higher-order predicate calculus.
\item The calculus used for the computation of such formula, and thus meaning assembly, is the $\lambda$-calculus with second order types, Girard's System-F.
\item The lexicon associate each word with a single main $\lambda$-term (as usual) and finitely many optional terms.
\item The infelicity of sentences (in case of a problem of selection) is evidenced by type mismatches during application.
\item The use of lexical information to resolve type mismatches is provided by the optional terms associated to the words, or main $\lambda$-terms, and guided by the types, rather than derived entirely from the type system.
\end{itemize}
In brief, our logic of assembly based on System-F and $TY_n$ (described in \cite{muskens-meaningandp}) comprises the following elements:

 \begin{itemize} 
\item 
Constants $\eee_i$ and $\ttt$, as well as any type variable $\alpha$ are types. 
\item 
Whenever $T$ is a type and $\alpha$ a type variable which may but need not occur in $T$,  $\Pi\alpha.\ T$ is a type. 
\item 
Whenever $T_1$ and $T_2$ are types, $T_1\fl T_2$ is a type as well.
\end{itemize}

Terms encode proofs of quantified propositional formulae: 
\begin{itemize} 
\item A variable  of type $T$ i.e. $x:T$ or  $x^{T}$  is a \emph{term}, and there are countably many variables of each type.
\item 
$(f\ \tau)$ is a term of type $U$ whenever $\tau:T$ and  $f:T\fl U$. 
\item 
$\lambda x^{T}.\ \tau$ is a term of type $T\fl U$ whenever $x:T$, 
and $\tau:U$.  
\item $\tau \{U\}$ is a term of type $T[U/\alpha]$
whenever $\tau:\Lambda \alpha.\ T$, and $U$ is a type. 
\item $\Lambda \alpha. \tau$ is a term of type $\Pi \alpha. T$
whenever $\alpha$ is a type variable, and  $\tau:T$ a term without any free occurrence of the type variable $\alpha$ in the type of a free variable of $\tau$.  
\end{itemize}

The latter restriction is the usual one on the proof rule for quantification in propositional logic: one should not conclude that $F[p]$ holds for any  proposition $p$
when assuming $G[p]$ --- i.e. having a free hypothesis of type $G[p]$. 

The reduction is defined by two schemes which are quite similar: 
\begin{itemize} 
\item $(\lambda x. \tau) u$ reduces to $\tau[u/x]$ (usual $\beta$ reduction). 
\item $(\Lambda \alpha. \tau) \{U\}$  reduces to $\tau[U/\alpha]$ (remember that $\alpha$ and $U$ are types). 
\end{itemize} 

As an example, in $Ty_n$ we need a first order quantifier per sort (or base type). 
Here, a single quantifier $\forall$ of type $\Pi \alpha.\ (\alpha \fl \ttt)\fl \ttt$ is sufficient. Indeed, this quantifier can be specialised to specific types, for instance to the base type $\zeta$, yielding $\forall \{\zeta\}: (\zeta \fl \ttt)\fl \ttt$, or even to properties  of $\zeta$ objects, which are of type $\zeta\fl\ttt$, yielding  
$\forall \{\zeta\fl\ttt\}: ((\zeta \fl \ttt) \fl \ttt)\fl \ttt$. 

\subsection{Application Revisited}

The types with $n$ sorts normally act as an implementation of selection restriction via application. The mechanisms of lexical semantics that convey additional meanings take place when the application encounter a type mismatch in a situation such as:
$$(P^{V\rightarrow W} x)\ \ \tau^U$$
In that case, the analysis looks for some optional term, available either through the lexical entry of $P$, $\tau$, or both, that can resolve the situation. It must be some $f^{U\rightarrow V}$, that is, the selection of this optional term is guided by the type system but not contained in type-derived rules.\par
If there is no suitable optional term available, the analysis ends with a type mismatch (or signals a missing operation, see section \ref{infelicity}). If there is at least one available, it is used to \emph{transform} the argument into a term that is suitable for the predicate, both yielding the expected type and providing a trace for the actual transformation in the meaning assembly phase. If there is more than one suitable transformation, multiple interpretations can occur.\par
In any case, after the transformation takes place, the type mismatch is corrected in the following way:
$$(P^{V\rightarrow W} (f^{U\rightarrow V} x) \ \  \tau^U$$
This leads to the semantic representation $P(f(\tau))$ after reduction.
\subsection{Example}

The optional terms are used to represent alternate lexical meanings induced from qualia, deverbals, or other alternations, which occur in the case of type mismatches in the usual meaning assembly. The following example is the lexical transformation from a city into the football club commonly associated, using the following (trimmed down) lexical entry:
$$\left < \mathrm{Liverpool}^{City}
 \  {; \  {{Id = \lambda x^{City} . x}}} \ ,
\ {{f_C^{City\rightarrow Club}}} \right >$$
Here, \emph{Liverpool} is considered as being primarily the proper name for an entity of type $City$, with a transformation that can provide something of type $Club$, that is, the football club commonly associated. Considering:
\begin{exe}
\ex \label{cup} Liverpool won the cup.
\end{exe}
For the sake of simplicity, this can be considered as a simple application of a predicate (\emph{won}) to an argument (\emph{Liverpool}). The predicate is the following:
$$\left <\lambda x^{Club} . (\mathrm{won}^{Club \rightarrow t} \ x)
 \  {; \  {{Id = \lambda x^{Club} . x}}} \right >$$
 
The application is the following:
$$\lambda x^{Club} . (\mathrm{won}^{(Club \rightarrow t} \ x) \ \mathrm{Liverpool}^{City}$$
The type mismatch can be resolved using the optional term $f_C$, contained in the lexical entry for \emph{Liverpool}, which gives the expected type. The application, using $f_C$ as a transformation, yields:
$$(\mathrm{won}^{Club \rightarrow t} \ (f_C^{City\rightarrow Club} \ \mathrm{Liverpool}^{City}))$$
The representation in predicate logic is simply $\mathrm{won}(f_C(\mathrm{Liverpool}))$.
\section{Recent Advances}
\subsection{Constraints}
One of the main problems with the usual treatment of polysemous words is to distinguish between felicitous and infelicitous co-predications. Consider:
\begin{exe}
\ex \begin{xlist}
	\ex The salmon was fast.
	\ex The salmon was delicious.
	\ex \label{salmon-s}*The salmon was fast and delicious.
	\ex \label{salmon-a}The salmon was lighting fast. It is delicious.
	\end{xlist}
\ex \begin{xlist}
	\ex Liverpool is a large city.
	\ex Liverpool voted Labour.
	\ex Liverpool won the cup.
	\ex \label{city-people}Liverpool is a large city and voted Labour.
	\ex \label{city-club}*Liverpool is a large city and won the cup.
	\end{xlist}
\end{exe}
Such data is subject to discussion and interpretation as to what is really felicitous or not. However, it has conducted us to provide a mechanism to distinguish between the word meanings (accessible by transformations) that are compatible with others, and those that are not.\par
We associate every optional term, every transformation and every term constructed in the process of the analysis with a \emph{degree of flexibility} that allow to provide \emph{constraints} associated with problematic transformations. Based on the behaviour seen above, transformation should at the least be distinguished between \emph{rigid} and \emph{flexible} ones. The flexibility of the terms is updated during the calculus and checked against the constraints in each application or conjunction.\par
As a first approximation, the difference is simply the following:
\begin{itemize}
\item Optional terms that are said to be \emph{flexible} ($F$) are compatible with everything else. There is no restriction to their behaviour.
\item Optional terms terms that are said to be \emph{rigid} ($R$) are only compatible with themselves. No other transformation may occur to a term they have modified, nor can they apply to a term that has been modified by any other.
\end{itemize}
The distinction is made within the lexicon. It is also necessary to keep track of this during the composition.\par
For instance, (\ref{city-club}) would be treated in the following way. Considering a slightly expended entry for \emph{Liverpool}:
$$\left < \mathrm{Liverpool}^{City}
 \  {; \  {{Id = \lambda x^{City} . x}\atop {F}} \ , \ {{f_P^{City\rightarrow People}}\atop {F}} \ , \  {{f_L^{City \rightarrow Location}} \atop {F}} \ ,
\ {{f_C^{City\rightarrow Club}} \atop {R}}} \right >$$
The identity transformation and the possibility to access either people or geographical location are flexible transformations (one can easily say (\ref{city-people})), the access to the football club is rigid. Simplifying slightly, we can summarise our example as a conjunction of two predicates, one pertaining to locations, the other of football clubs:
$$\left <\lambda x^{Location} . (\mathrm{large}^{Location \rightarrow t} \ x)
 \  {; \  {{Id = \lambda x^{Location} . x}\atop {F}}} \right >$$

$$\left <\lambda x^{Club} . (\mathrm{win}^{Club \rightarrow t} \ x)
 \  {; \  {{Id = \lambda x^{Club} . x}\atop {F}}} \right >$$

When parsing flexibility, it is impossible to apply simultaneously the flexible and rigid transformations:
\begin{center}
\leaf{Liverpool}
\branch{1}{$(f_L \ \mathrm{Liverpool})$ : F}
\branch{1}{$(\mathrm{large} \ \mathrm{Liverpool})$ : F}
\leaf{Liverpool}
\branch{1}{$(f_C \ \mathrm{Liverpool})$ : R}
\branch{1}{$(\mathrm{win} \ (f_C \ \mathrm{Liverpool}))$ : R}
\branch{2}{$\wedge$ : $\varnothing$}
\tree \end{center}
On the other and, a \emph{correct} co-predication such as (\ref{city-people}) is made possible by the use of flexible optional terms as transformations. With a predicate pertaining to \emph{People}, such as:
$$\left <\lambda x^{People} . (\mathrm{lively}^{People \rightarrow t} \ x)
 \  {; \  {{Id = \lambda x^{People} . x}\atop {F}}} \right >$$
The conjunction is acceptable because only flexible transformations are used:
\begin{center}
\leaf{Liverpool}
\branch{1}{$(f_L \ \mathrm{Liverpool})$ : F}
\branch{1}{$(\mathrm{large} \ \mathrm{Liverpool})$ : F}
\leaf{Liverpool}
\branch{1}{$(f_P \ \mathrm{Liverpool})$ : F}
\branch{1}{$(\mathrm{lively} \ (f_P \ \mathrm{Liverpool}))$ : F}
\branch{2}{$\wedge$ : F}
\tree \end{center}
The resolution of a type mismatch in a simple application is trivial. Co-predication, however, requires a conjunction between two predicates having different types as targets. This is accomplished by using a polymorphic version of \emph{and}:
$$\Land =
\Lambda \alpha \Lambda \beta
\lambda P^{\alpha \fl \ttt} \lambda Q^{\beta\fl \ttt} 
 \Lambda \xi \lambda x^\xi 
 \lambda f^{\xi\fl\alpha} \lambda g^{\xi\fl\beta}.\ 
(\textrm{and}^{\ttt\fl\ttt\fl\ttt} \ (P \ (f \ x)) (Q \ (g \  x))) 
$$
This operator takes the conjunction of two arbitrary predicates $P$ and $Q$ taking two different types $\alpha$ and $\beta$ as their argument, and provides a guide for applying two possibly different optional terms as transformations for any argument of arbitrary type $\xi$. The application becomes the following:

$$(\Lambda \alpha \Lambda \beta
\lambda P^{\alpha \fl \ttt} \lambda Q^{\beta\fl \ttt} 
 \Lambda \xi \lambda x^\xi 
 \lambda f^{\xi\fl\alpha} \lambda g^{\xi\fl\beta}.\ 
(\textrm{and}^{\ttt\fl\ttt\fl\ttt} \ (P \ (f \ x)) (Q \ (g \  x)))) \ \{Location\} \ \{People\}$$
$$/ \ \mathrm{large}^{Location\rightarrow \ttt}
 \ \mathrm{lively}^{People\rightarrow \ttt} \ \{City\} \ \mathrm{Liverpool}^{City} \ f_L^{City\rightarrow Location} \ f_P^{City\rightarrow People}$$
This reduces to:
$$(\mathrm{and}^{\ttt \rightarrow \ttt \rightarrow \ttt} \ (\mathrm{large}^{Location\rightarrow \ttt} \ (f_L^{City\rightarrow Location} \ \mathrm{Liverpool}^{City}))\ (\mathrm{lively}^{People\rightarrow\ttt}\ (f_P^{City\rightarrow People}\ \mathrm{Liverpool}^{City}))$$
and yields the wanted representation in predicate calculus, of
$$\mathrm{and}(\mathrm{large}(f_L(\mathrm{Liverpool})), \mathrm{lively}(f_P(\mathrm{Liverpool})))$$

\subsection{Syntax-Driven Constraint Relaxing}
However, some rigid transformations exhibit different behaviours when used in a longer syntactic unit than a sentence. Anaphoric reference can make sentences such as (\ref{salmon-a}) felicitous, while (\ref{salmon-s}) is not.\par
A first refinement would then be to distinguish between transformations that are \emph{flexible} (of degree 1), \emph{semi-flexible} (of degree 2), and \emph{rigid} (of degree 3). The semi-flexible transformations would have the behaviour of rigid transformations, but their use would be destroyed by syntactic constructs such as anaphoric reference.
\subsection{Articles and Quantifiers}
\label{articles}
The usual view in Montague Grammar (as seen in section \ref{montague}) is that common nouns such as \emph{book} are accurately modelled as predicates (the set of properties that means to be a book), and that articles are quantifiers that can point to individuals satisfying such predicates (and thus \emph{a book} is an entity). The usual treatment, using existential, universal or generalised quantifiers is questionable in the standard Montagovian theory, and becomes very problematic when used with $n$ sorts.\par
Hilbert's operators provide an interesting alternative to generalised quantifiers. Described in \cite{hilbert-grundlagen}, $\epsilon$ and $\iota$ can be used to represent some aspects of determiners, and have been used in this fashion by von Heusinger in \cite{vheusinger-epsilon} and \cite{vheusinger-choice}. They can be used in order to model the existential quantification and dynamically linking.\par
In Montague Grammar, $\epsilon$ is of type $(\eee\rightarrow \ttt)\rightarrow \eee$, making it a good choice for the semantics of the indefinite article \emph{a}: if \emph{book} is of type $(\eee\rightarrow\ttt)$, \emph{a book} is of type $\eee$. But our framework, based on principles borrowed from the Generative Lexicon, uses many types and \emph{book} would be typed as $(Readable\rightarrow \ttt)$, with $Readable$ being an ontological type common to most printed works, and the work \emph{book} providing optional terms in order to access the physical object associated and the informational content.\par
Our indefinite determiner is modelled as a polymorphic, many-sorted version of Hilbert's $\epsilon$: a constant $\epsilon$ of type $\Lambda\alpha . (\alpha\rightarrow\ttt)\rightarrow\alpha$. As the original operator, given a predicate with some arbitrary target types, it provides an individual of that type.\par
However, a different approach argued by Luo is that common nouns simply provide a base type; see section \ref{types} for discussion.
\newpage
\subsection{Linear Formulation for more Flexible Constraints}
The incompatibility of some facets is, up to now, accounted for by the rigidity of some transformations. 
Such transformations are exclusive: they exclude any other transformation. 
This is somehow too schematic and it is possible that the actual incompatibilities are more subtle than this. 
Assuming that there are four transformations associated with a word, $f,g,h,k$ it is \emph{a priori} possible 
--- although we do not have examples of this yet --- that the compatible subsets are $\{k\},\{f,g\},\{g,h\},\{h,f\}$ but not $\{f,g,h\}$, or even more sophisticated configurations.\par
Such explicit subsets might be formulated using a single transformation which produces the compatible subsets. 
If these were types, linear logic  would easily  express these possibilities, e.g. with $!k\& (!f\otimes !g) \& (!g\otimes !h)\ \& (!h\otimes !f)$ or the alternative using $!a\otimes !b\equiv !(a\& b)$ 
--- second order intuitionnistic linear logic faithfully encodes system-F, as studied  in  \cite{Ret87}.  We are presently studying whether the proof terms of 
\cite{DBLP:conf/tlca/BentonBPH93} for intutitionnistic linear logic could be used in the same manner, with transformations as constants, i.e. as black boxes of the corresponding types.\par
This seems to be a promising direction for handling subtle incompatibilities --- and at the same time we are looking for linguistic data with such tricky incompatibilities. 
\section{Discussions}
\subsection{Transitivity}
The issue of the transitivity of transformations is a different issue that of constraints that can apply to them. Consider:
\begin{exe}
\ex \begin{xlist}
\ex \label{club1}Liverpool won the cup.
\ex \label{club2}It had hired several oversea players.
\ex \label{club3}This caused it to go bankrupt.
\ex \label{club4}*Liverpool was thus forced to resign.\end{xlist}
\end{exe}
In (\ref{club1}), \emph{Liverpool} is used in the sense of the football club, via a first transformation. The \emph{club} itself receives a second transformation in (\ref{club2}) (despite the first transformation being rigid: this is not a case of conflict between several senses, but of sequential changes), that makes accessible some unnamed person responsible for the management of the club. This can carry for a bit, but there are clearly limits, as seen in (\ref{club4}).\par
We can remark that idioms (that is, linguistic differences) can make some transformations possible:
\begin{exe}
\ex \begin{xlist}
\ex \label{carT}My car has a punctured tyre.
\ex \label{carP}*My car is punctured.
\ex \label{voiture}Ma voiture est crevée.\end{xlist}
\end{exe}
If (\ref{carT}) is possible, (\ref{carP}) should be under the assumptions of \cite{gen-lexicon} that a \emph{tyre} is an important constitutive part of a \emph{car}. It is not, but the French (\ref{voiture}) is, which is a literal translation of (\ref{carP}). This illustrate the importance of making the lexical transformations on a linguistical rather than ontological basis.\par
However, consider:
\begin{exe}
\ex \begin{xlist}
\ex \label{carI}My car has four injectors.
\ex \label{cloggedI}One of them is clogged.
\ex \label{cloggedC}? My car is clogged.
\end{xlist}\end{exe}
(\ref{carI}) followed by (\ref{cloggedI}) is felicitous. However, in the context of (\ref{carI}), (\ref{cloggedC}) can be understood, if a bit strange, but it is clearly infelicitous by itself.\par
Keeping track of successive transformations and what can be ultimately derived from a word is thus as necessary as the evaluation of what transformations are possible simultaneously.

\subsection{Types, Predicates and Granularity}
\label{types}
There are two closely related issues with regard to the design of the type system. The first is wether nouns should be represented as being individuals of some base type or predicates, the second is the definition of the set of base types. There are several possibilities for the latter:
\begin{itemize}
\item The only sort is $\eee$. This preclude any type-based lexical treatment.
\item There is a finite, small number of sorts that serve as base types, derived from ontological properties, as envisioned by \cite{gen-lexicon} and \cite{asher-webofwords} (e.g. $Physical$, $Information$, $Agent$, $Artifact$\ldots~ All told, two or three dozen at most).
\item There is a finite and relatively small number of sorts that are defined by the common restrictions of selection gathered from the grammar of a language ($animate$, $human$, $male$\ldots~about 200 in total).
\item There is a finite but large number of base types: one for every common noun, and many other derivable $\Sigma$-types (\cite{luo-types}).
\item Every possible formula with a free variable is a type. This could lead to a circular definition (as types are used to define formulae).
\end{itemize}
For some of these sets of base types, having nouns as individuals rather than predicates is a sound approach, specifically in the case where every noun defines its own type. Our view is that for some finite and reasonable number of base types, the two approaches can be non-contradictory.\par

It is possible to envision both possibilities simultaneously, by associating to every base type $\alpha$ the predicate \emph{to be of type $\alpha$}, written $\hat{\alpha}$, which is of type $\eee \rightarrow \ttt$. In that case, it is possible to quantify over individuals of a given type using this construct and the polymorphic $\epsilon$ given in section \ref{articles}.
\subsection{Locality and Contextuality}
There are many cases where differences between social contexts or speakers can result in different lexica. The most obvious example is in fictional narratives such as mythology, fables, fairy tales, fantasy or science-fiction, where a great part of the vocabulary is attached to the narrative. Some of these works make use of standard vocabulary but change some of the settings (for instance, children's literature makes extensive use of animals as characters, effectively changing their type to $Agent$), some introduce specific words either by amalgamation (\emph{direwolf}, \emph{lightsaber}, etc.) or similarity with historical words (\emph{serjeant}), and some use words that are not related to the reader's lexicon and have him infer their meaning purely from context (Tolkien's Quenda and Sindar, as most readers will not get around to perusing an Elvish dictionary).\par
Each of those literary contexts modify the lexicon as part as the normal reading process of the work associated.
We believe that Fantasy, SF and tales can be modelled as having a slightly \emph{expanded} lexicon, with additional words available and some additional senses available for some words. There are many words that are not used, but they are ignored (as in any other case where the full lexicon is not used for a particular text).\par
\newpage
The idea here is that additions to the lexicon can be made locally in the following cases:
\begin{itemize}
\item There is a known difference in the literary genre.
\item There is an apparent difference in the vocabulary used by different speakers in a dialogue.
\item There is a social setting that provides additional terms, such as those used in a trade \emph{jargon}.
\end{itemize}

\paragraph{Examples} -- all of the following are from G. R. Martin's \emph{A Dance with Dragons}.\par
\begin{itemize}
\item The point of view is of particular importance. Even outside of dialogue, where the narration is third-person, a specific agent is used, explicitly (Robbin Hobb, George Martin\ldots) or implicitly, for providing atmospheric effects and involving descriptions:
\begin{exe}
\ex But the air was sharp and cold and full of fear.\end{exe}

\item The text will often use additional words, that do not belong in any language. Here is an explicit example, that include the necessary information for the reader's lexicon:
\begin{exe}
\ex Their driver awaited them beside his \emph{hathay}. In Westeros, it might have been called an oxcart, though it [...] lacked an ox. The \emph{hathay} was pulled by a dwarf elephant[...]\end{exe}

\item However, the mechanisms of lexical semantics apply equally as well in such a setting. For instance, organisations such as banks continue to have their multiple aspects available, including an unnamed human leader:
\begin{exe}
\ex We who serve the Iron Bank face death full as often as you who serve the Iron Throne.\end{exe}
\end{itemize}
In each case, a slightly expanded and differentiated lexicon would be needed.
%
%
%
%
%

\subsection{Infelicitous statements}
\label{infelicity}
In the view of such examples and many others where specific senses are associated to words in a technical jargon (skeumorphic metaphors in the computer world such as \emph{file}, \emph{document}, \emph{directory}, \emph{window}\ldots, for instance, became mainstream with the technology), there is a danger in taking the view that infelicitous sentences are simply wrongly formed. In our view, the analyser should not necessarily reject sentences that are infelicitous because of a type mismatch with no suitable transformation available, but instead indicate that there must exist such a transformation that is missing from the lexicon, and try to infer its characteristics.

\subsection{Other approaches}
There have been several other approaches to the modelling of a complete framework for the Generative Lexicon. However, the first ones (including the original formulation by James Pustejovsky and early work by Nicholas Asher) neglected large parts of compositional semantics based on the syntactic structure, including determiners, quantifiers, plurals\ldots~Our own approach, as well as the more recent others, have focused on the integration of this aspect.
\newpage
\paragraph{Asher} -- The first formulation was arguably \cite{pustejovsky-metaphysics}, which was continued by Nicholas Asher in several works that culminate in \cite{asher-webofwords}. The most advanced of those approaches use \emph{type presuppositions} as a guide for the computation of meaning. The formal aspect of the computation of complex types, however, is still a problem, as is their interpretation in term of categorical fiber-products.\par
\paragraph{Cooper} -- In \cite{cooper-codygeqlic}, Robin Cooper proposes a complete model that uses record types in Per Martin-Löf's type theory, as well as generalised dynamic quantifiers. We are continuing discussions regarding this approach and possible convergences.\par
\paragraph{Luo} -- Luo Zhaohui has been working with Nicholas Asher and has proposed a slightly different logic based on Unified Type Theory, in \cite{luo-dot}, and recently \cite{luo-contextual}. His principles are to consider words as types and transformations as coercions in a paradigm of coercitive subtyping. Discussions have proven interesting with regards to many aspects of the logic and type system.
\section*{Conclusion}
The work presented here is still in progress. We are still looking for a satisfactory implementation of constraints using types in linear logic, and for an account of the limits of transitivity. There are three main areas for future research:
\begin{itemize}
\item Exploration of additional phenomena, including nonce words and hypostasis, the relativity of the lexicon to social contexts and agents. Such research is currently undertaken by Cooper.
\item Fine-tuning the formal framework, including a formal approach of quantification and syntax-driven composition, as Asher and Luo are doing.
\item Implementing and testing a complete framework.
\end{itemize}
We are making some progress in all three of these areas. We are investigating alternate literary universes. We have made some real progress from the original formulation on the formal compositional model, with accepted publications forthcoming on the use of Hilbert's $\epsilon$ for indefinite articles and on plurals (\cite{moot-plurals}). We also have a partial implementation in $\lambda$-DRT, using Richard Moot's Grail analyser for syntax and semantics (\cite{moot-grail}), including applications to the virtual traveler problem (\cite{moot-traveler}), among others. Much is still missing, however, especially work on the automatic acquisition of rich, well-typed lexica. Progress in that direction remains dependent on further formalisation on the granularity of the type system, and we are optimistic.

\bibliographystyle{plain}
\bibliography{sources}

\begin{thebibliography}{10}

\bibitem{asher-fundamenta}
Nicholas Asher.
\newblock {A Type Driven Theory of Predication with Complex Types}.
\newblock {\em {Fundamenta Informatic\ae{}}}, 84(2):151--183, 2008.

\bibitem{asher-webofwords}
Nicholas Asher.
\newblock {\em {Lexical Meaning in Context: a Web of Words}}.
\newblock {Cambridge University Press}, March 2011.

\bibitem{mery-jolli}
Christian Bassac, Bruno Mery, and Christian Retoré.
\newblock {Towards a Type-Theoretical Account of Lexical Semantics}.
\newblock {\em {Journal of Language, Logic, and Information}}, 19(2), 2010.

\bibitem{DBLP:conf/tlca/BentonBPH93}
P.~N. Benton, Gavin~M. Bierman, Valeria de~Paiva, and Martin Hyland.
\newblock A term calculus for intuitionistic linear logic.
\newblock In Marc Bezem and Jan~Friso Groote, editors, {\em TLCA}, volume 664
  of {\em Lecture Notes in Computer Science}, pages 75--90. Springer, 1993.

\bibitem{cooper-codygeqlic}
Robin Cooper.
\newblock {Copredication, dynamic generalized quantification and lexical
  innovation by coercion}.
\newblock In {\em {Fourth International Workshop on Generative Approaches to
  the Lexicon}}, 2007.

\bibitem{copestake-lexical}
Ann Copestake and Ted Briscoe.
\newblock Lexical operations in a unification based framework.
\newblock In {\em ACL SIGLEX Workshop on Lexical Semantics and Knowledge
  Representation}, 1991.

\bibitem{vheusinger-epsilon}
U.~Egli and K.~von Heusinger.
\newblock The epsilon operator and e-types pronouns.
\newblock {\em Lexical Knowledge in the Organization of Language}, pages
  121--141, 1995.

\bibitem{hilbert-grundlagen}
D.~Hilbert and P.~Bernays.
\newblock {\em {Grundlagen der Matematik. Bd. 2}}.
\newblock Springer, 1939.

\bibitem{luo-contextual}
Zhaohui Luo.
\newblock Contextual analysis of word meanings in type-theoretical semantics.
\newblock In {\em {Pogodalla and Prost}}, pages 159--174. 2011.

\bibitem{luo-types}
Zhaohui Luo.
\newblock Common nouns as types.
\newblock In {\em {BÃ©chet and Dikovsky}}, pages 173--185. 2012.

\bibitem{mery-jsm}
Bruno Mery.
\newblock {Compositionality and the Lexicon (Lexique organisé pour la
  composition sémantique)}.
\newblock In {\em {Journées Sémantique et Modélisation}}, Paris, France, 2009.

\bibitem{mery-montagl}
Bruno Mery, Christian Bassac, and Christian Retor{\'e}.
\newblock A montagovian generative lexicon.
\newblock In {\em Formal Grammar}, 2007.

\bibitem{mery-ndttg}
Bruno Mery, Christian Bassac, and Christian Retor{\'e}.
\newblock A montague-based model of generative lexical semantics.
\newblock In Reinhard Muskens, editor, {\em New Directions in Type Theoretic
  Grammars}. ESSLLI, Foundation of Logic, Language and Information, 2007.

\bibitem{montague-proper}
Richard Montague.
\newblock {The proper treatment of quantification in ordinary English}.
\newblock In R.~H. Thomson, editor, {\em {Formal Philosophy}}, pages 188--221.
  {Yale University Press}, {New Haven Connecticut}, 1974.

\bibitem{moot-grail}
Richard Moot.
\newblock {Wide-coverage French syntax and semantics using Grail}.
\newblock In {\em {Proceedings of Traitement Automatique des Langues Naturelles
  (TALN}}, Montreal, 2010.

\bibitem{moot-plurals}
Richard Moot and Christian RetorÃ©.
\newblock Second order lambda calculus for meaning assembly: on the logical
  syntax of plurals.
\newblock In {\em Coconat}, {Tilburg, Netherlands}, 2011.

\bibitem{moravcik-article}
Julius~M. Moravcsik.
\newblock How do words get their meanings ?
\newblock {\em The Journal of Philosophy}, LXXVIII(1), January 1982.

\bibitem{muskens-meaningandp}
Reinhard Muskens.
\newblock {Meaning and Partiality}.
\newblock In Robin Cooper and Maarten de~Rijke, editors, {\em {Studies in
  Logic, Langage and Information}}. {CSLI}, 1996.

\bibitem{pustejovsky-type}
J.~Pustejovsky.
\newblock Type construction and the logic of concepts, 2001.

\bibitem{pust-genlex-article}
James Pustejovsky.
\newblock The generative lexicon.
\newblock {\em Computational Linguistics}, 17(4):409--441, 1991.

\bibitem{gen-lexicon}
James Pustejovsky.
\newblock {\em {The Generative Lexicon}}.
\newblock {MIT Press}, 1995.

\bibitem{pustejovsky-metaphysics}
James Pustejovsky and Nicolas Asher.
\newblock {The Metaphysics of Words in Context}.
\newblock {\em {Objectual attitudes, Linguistics and Philosophy}}, 23:141--183,
  February 2000.

\bibitem{moot-traveler}
L.~PrÃ©vot R.~Moot and C.~RetorÃ©.
\newblock {Un calcul de termes typÃ©s pour la pragmatique lexicale -- chemins
  et voyageurs fictifs dans un corpus de rÃ©cits de voyages}.
\newblock In {\em {Traitement Automatique du Langage Naturel -- TALN 2011}},
  pages 161--166, {Montpellier, France}, 2011.

\bibitem{real-deverbal}
L.-M. Real-Coelho and C.~RetorÃ©.
\newblock A generative montagovian lexicon for polysemous deverbal nouns.
\newblock In {\em 4th World Congress and School on Universal Logic -- Workshop
  on Logic and Linguistics}, Rio de Janeiro, 2013.

\bibitem{Ret87}
C.~Retor{\'e}, Christian.
\newblock Le syst{\`e}me {F} en logique lin{\'e}aire.
\newblock {M}{\'e}moire de {D}.{E}.{A}. (dir.: J.-y. girard), Universit{\'e}
  Paris 7, 1987.

\bibitem{vheusinger-choice}
K.~von Heusinger.
\newblock Choice functions and the anaphoric semantics of definite nps.
\newblock {\em Research on Language and Computation}, 2:309--329, 2004.

\bibitem{luo-dot}
T.~Xue and Z.~Luo.
\newblock Dot-types and their implementation.
\newblock In {\em {BÃ©chet and Dikovsky (2012), pages 234-249}}, 2012.

\end{thebibliography}

\end{document}